\pdfoutput=1
\documentclass[11pt]{article}

\usepackage[preprint]{coling}

\usepackage{times}
\usepackage{latexsym}

\usepackage[T1]{fontenc}
\usepackage[utf8]{inputenc}

\usepackage{microtype}

\usepackage{inconsolata}

\usepackage{graphicx}
\usepackage{subcaption}
\usepackage{xcolor}
\usepackage{float}
\usepackage{stfloats}

\usepackage{listings}

\title{Beyond Text: Optimizing RAG with Multimodal Inputs for Industrial Applications}

\author{
 \textbf{Monica Riedler\textsuperscript{1,2}},
 \textbf{Stefan Langer\textsuperscript{1,2}}
\\
\\
 \textsuperscript{1}Center for Information and Language Processing, LMU Munich,
 \\
 \textsuperscript{2}Siemens AG
\\
    \texttt{\href{mailto:monica.riedler@campus.lmu.de}{monica.riedler@campus.lmu.de}},
    \texttt{\href{mailto:stefan.langer@cis.lmu.de}{stefan.langer@cis.lmu.de}}
}

\begin{document}
\maketitle
\begin{abstract}
Large Language Models (LLMs) have demonstrated impressive capabilities in answering questions, but they lack domain-specific knowledge and are prone to hallucinations. 
Retrieval Augmented Generation (RAG) is one approach to address these challenges, while multimodal models are emerging as promising AI assistants for processing both text and images.
In this paper we describe a series of experiments aimed at determining how to best integrate multimodal models into RAG systems for the industrial domain. The purpose of the experiments is to determine whether including images alongside text from documents within the industrial domain increases RAG performance and to find the optimal configuration for such a multimodal RAG system.
Our experiments include two approaches for image processing and retrieval, as well as two LLMs (GPT4-Vision and LLaVA) for answer synthesis. These image processing strategies involve the use of multimodal embeddings and the generation of textual summaries from images.
We evaluate our experiments with an LLM-as-a-Judge approach. Our results reveal that multimodal RAG can outperform
single-modality RAG settings, although image retrieval poses a greater challenge than text retrieval. Additionally, leveraging textual summaries from images presents a more promising approach compared to the use of multimodal embeddings, providing more opportunities for future advancements.
\end{abstract}

\section{Introduction}

The release of Large Language Models (LLMs), such as Llama3 \citep{MetaLLaMA3} and GPT-4 \citep{gpt4-_technical_report}, has significantly advanced the field of Natural Language Processing (NLP), enabling a wide range of applications, including automated content generation and conversational agents. However, LLMs often still lack domain-specific knowledge and are prone to hallucinations \citep{domain_specific, rawte-etal-2023-hallucination}. Retrieval Augmented Generation (RAG) addresses these limitations by combining document retrieval with generative language models.

Recently, Multimodal Large Language Models (MLLMs) have emerged, extending LLM capabilities to include modalities like images and videos \citep{zhang-etal-2024-mm, survey_multimodal_llms}. This development holds significant potential for industrial settings such as manufacturing, engineering, and maintenance, where documents like manuals, software guides, and product brochures frequently combine complex technical text with detailed visuals, such as diagrams, schematics and screenshots. This combination of modalities makes the industrial domain particularly challenging for AI systems, as they must accurately interpret both textual and visual information to provide meaningful insights.

While extensive research has been conducted on text-only RAG systems and their optimization \citep{Gao2023rag_survey, siriwardhana-etal-2023-improving}, the application of multimodal RAG to the industrial domain is less documented in academic literature. 
Existing examples mainly target general-domain datasets \citep{chen-etal-2022-murag, lin-byrne-2022-ravqa} and medical applications \citep{sun2024fact, xia2024rule, zhu2024emerge}.

In our paper, we explore the integration of multimodal models into RAG systems for the industrial domain. Specifically, we investigate whether incorporating images alongside text enhances RAG performance and we identify optimal configurations for such systems. We use two MLLMs, GPT-4Vision \citep{2023GPT4VisionSC} and LLaVA \citep{improved_llava}, for answer synthesis and evaluate two image processing strategies: multimodal embeddings and textual summaries from images.

Our research addresses two primary questions: (1) Does the inclusion of both text and images improve the performance of RAG systems in the industrial domain compared to single-modality RAG? (2) How can the performance of multimodal RAG be optimized for this domain? To answer these questions, we compare the performance of single-modality (text-only or image-only) and multimodal (text and image) RAG systems using a set of 100 domain-specific questions. Additionally, we explore various retrieval methods to optimize the performance of the multimodal RAG pipelines.

Our paper makes the following contributions:
\begin{itemize}
    \item We integrate multimodal models into RAG systems for the industrial domain, demonstrating that multimodal RAG can outperform single-modality RAG.
    \item We compare multimodal embeddings and image summaries for image processing, employing GPT-4V and LLaVA for answer synthesis, and find that image summaries offer greater flexibility and potential for advancement.
\end{itemize}

\section{Related Work}
  
\paragraph{Multimodal LLMs} In recent years, LLMs have shown emergent abilities such as in-context learning, instruction following, and chain-of-thought reasoning \citep{gpt3, instructgpt, cot_prompting}, making them suitable for various NLP tasks. Multimodal LLMs extend these capabilities to understand and generate multiple modalities, enabling AI assistants to process text, images, videos, and audio. Notable examples include the integration of pre-trained unimodal models into multimodal systems, which consist of a modality encoder, a pre-trained LLM, and a modality generator, connected by projectors to transform inputs and outputs across different modalities \citep{zhang-etal-2024-mm, survey_multimodal_llms}.

\paragraph{Retrieval Augmented Generation} RAG has emerged as an effective approach to address the limitations of LLMs in domain-specific question answering. By combining an LLM for answer generation with an external vector database accessed via a retriever, RAG has been effectively applied to various tasks, including question answering, fact verification, and question generation, achieving state-of-the-art results in open-domain question answering \citep{lewis2020rag, guu, izacard-grave-2021-leveraging, pmlr-v162-borgeaud22a_retro}. 
Key optimizations to the initial approaches are extensively described by \citet{Gao2023rag_survey}. Despite significant advancements, challenges such as retrieval quality \citep{ma-etal-2023-query_rewriting, query_expansion}, the reliability of the generation component \citep{wu2024clashevalquantifyingtugofwarllms, niu-etal-2024-ragtruth}, and RAG robustness remain active areas of research \citep{rag_robustness}.

\paragraph{Multimodal RAG}  
\citet{chen-etal-2022-murag} introduced MuRAG, the first multimodal Retrieval-Augmented Transformer, which enhances model capabilities using an external non-parametric multimodal memory. An alternative approach by \citet{lin-byrne-2022-ravqa} involves transforming images into textual representations through OCR, image captioning, and object detection, followed by dense passage retrieval \citep{karpukhin-etal-2020-dense}. 
Other works apply multimodal RAG scenarios to medical and healthcare applications \citep{sun2024fact, xia2024rule, zhu2024emerge}, highlighting the potential of leveraging images as additional context.

\paragraph{RAG Evaluation} 
Evaluating RAG systems involves assessing both retrieval and generation components. Frameworks like RAGAs \citep{es-etal-2024-ragas} incorporate metrics such as Faithfulness, Answer Relevance, and Context Relevance to holistically assess RAG performance. Automated benchmarking methods, using datasets like TruthfulQA \citep{lin-etal-2022-truthfulqa} and MMLU \citep{hendrycks2021mmlu}, also evaluate specific RAG capabilities. For our experiments, we adopt an LLM-as-a-Judge approach \citep{chiang2024chatbot}, where LLMs evaluate their own generated outputs.
While scalable and often aligning with human evaluations, automated methods remain approximations. Human annotations are the gold standard for accuracy, especially in domain-specific tasks, but are costly and time-intensive, making automated methods a practical alternative. For multimodal evaluation, \citet{Zhang2023GPT4V_evaluator} showed GPT-4V’s effectiveness in vision-language tasks, aligning well with human assessments.

\section{Approach}

\subsection{Data}

Due to the lack of annotated context-question-answer triplets in the industrial domain, and even more so for a multimodal setting requiring quadruples of text context, image context, question, and answer, we created a manually annotated dataset\footnote{Unfortunately we cannot release this dataset due to copyright concerns.}.

We used 20 PDF documents from the industrial domain, such as manuals and software documentation for devices like programmable controllers, circuit breakers, and robots. Text and images were extracted from these documents, resulting in 8540 text chunks (with an average length of 225 words per chunk) and 8377 images, aligned by page to maintain contextual accuracy.

To create the RAG test set, we manually annotated 100 question-answer pairs. Each annotation includes a question, reference answer, and page number used to retrieve both text and image context from the corresponding page, forming multimodal quadruples. The questions were designed to address typical industrial tasks, such as operational procedures, device configurations, and troubleshooting, where visual context plays a key role.
The annotation process involved two annotators: one with basic domain knowledge and the other with extensive industrial experience. 
An example quadruple, along with further details on the extraction and annotation, are provided in Appendix \ref{sec:dataset_creation}.

\subsection{Experiments}

In this section, we outline the experiments we conducted to investigate two primary questions: (1) Does using both text and image modalities improve performance? (2) What is the optimal configuration for a multimodal RAG system in the industrial domain? We categorize the experiments into three RAG settings: Text-Only RAG, Image-Only RAG, and Multimodal RAG. Additionally, we implemented two reference settings for comparison: (1) a Baseline, where we feed questions directly to an LLM without retrieval, and (2) a Gold Standard Context setting, which serves as an upper bound.
We used a prompt to instruct the model to answer based on the retrieved context (Figure \ref{fig:qa_template.pdf} in Appendix).
To ensure consistency across experiments, we ran all settings with both GPT-4V and LLaVA, including the text-only settings, which do not require multimodal content processing.
We report implementation details on vector databases, retrieval, model versions, and hyperparameters in Appendix \ref{sec:implementation_details}.
  
\subsubsection{Baseline}
  
Our baseline consists in feeding questions from the test set directly to the LLM, without any retrieval step. 
This allowed us to test the LLM's internal knowledge and gain insight into its performance on domain-specific industrial questions.
  
\subsubsection{Text-Only RAG}
  
In the Text-Only RAG setting, we used only the texts  extracted from the PDF collection. We embedded the text chunks using OpenAI’s text-embedding-3-small\footnote{\url{https://platform.openai.com/docs/guides/embeddings}} model and stored them in a vector store. We then performed a similarity search on the vector store for each embedded question to retrieve the most relevant texts, which were concatenated with the user query and passed to the multimodal LLM for answer generation. This setup allowed us to evaluate the performance of text-based retrieval and answer synthesis.

\subsubsection{Image-Only RAG}

\begin{figure*}[ht!]
    \centering
    \begin{subfigure}{0.65\textwidth}
        \centering
        \includegraphics[width=\textwidth]{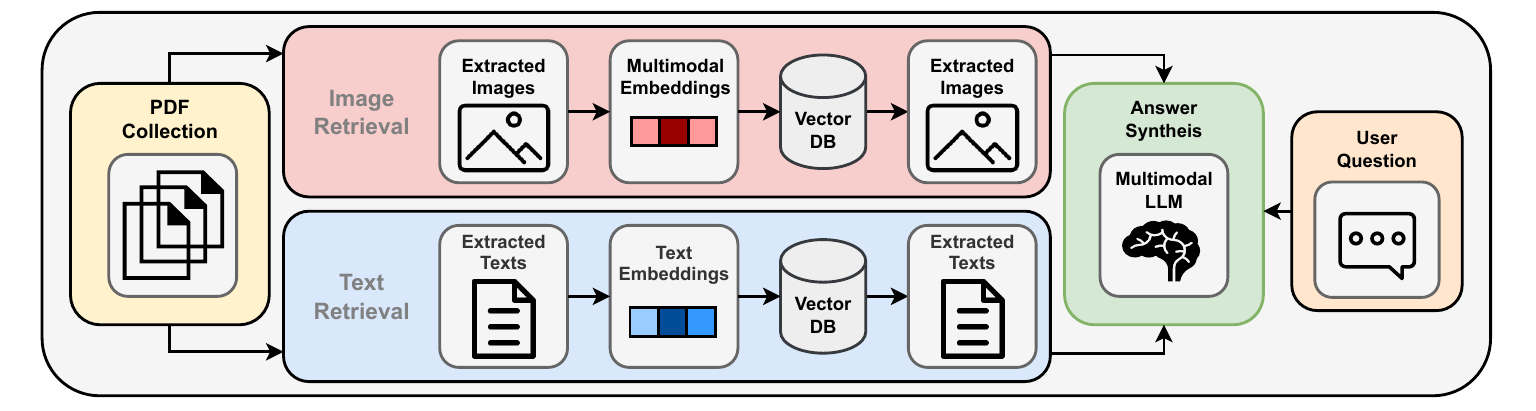}
        \caption{Multimodal RAG with Multimodal Embeddings and Separate Vector Stores.}
    \end{subfigure}
    \begin{subfigure}{0.80\textwidth}
        \centering
        \includegraphics[width=\textwidth]{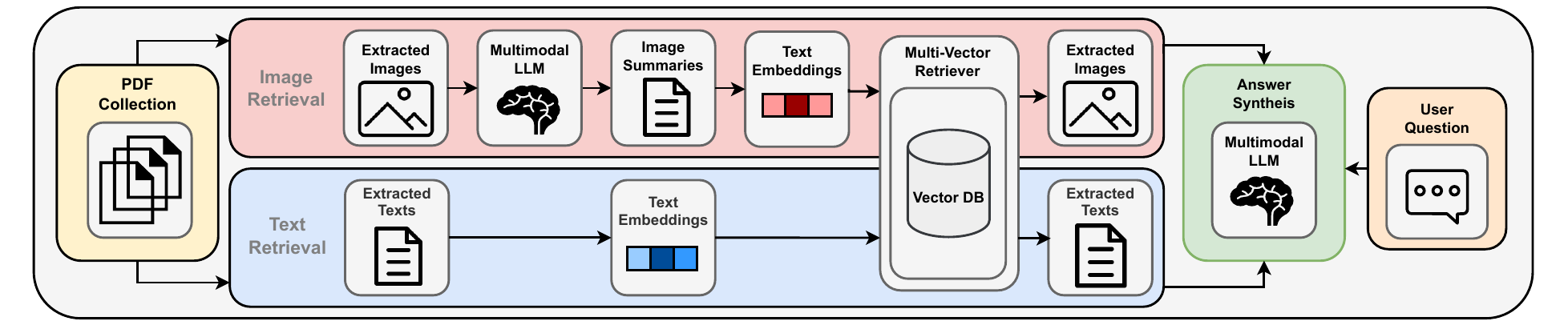}
        \caption{Multimodal RAG with Image Summaries and Combined Vector Store.}
    \end{subfigure}
    \caption{Overall architecture of our proposed multimodal RAG pipelines. For the Text-Only RAG, we only use the \textcolor{blue}{Text Retrieval} component. Conversely, in the Image-Only RAG, we only employ the \textcolor{red}{Image Retrieval} component, either with multimodal embeddings or image summaries.}

    \label{fig:mm_rag_overview}
\end{figure*}
  
In this setting, we only used images from the PDF documents. For image retrieval, we explored two distinct approaches:

\paragraph{Multimodal Embeddings}

We used CLIP \citep{clip-v139-radford21a} to jointly embed both images and questions. CLIP was selected for its ability to align image and text modalities in a shared embedding space, which allows to easily compute similarities between different types of data. This alignment is crucial for multimodal retrieval tasks, where understanding the relationship between image content and textual queries is key. We stored the obtained embeddings in a vector store, and performed a similarity search to retrieve the most relevant images based on the embedded query.

\paragraph{Text Embeddings From Image Summaries}
  
We summarized the images into text using a multimodal LLM (see Figure \ref{fig:summarization_template.pdf} in Appendix) and then embedded these summaries using text-embedding-3-small. We employed LangChain’s Multi-Vector Retriever\footnote{\url{https://python.langchain.com/v0.2/docs/how_to/multi_vector/}}, which allows to decouple the retrieval and generation sources. Summaries were stored in a vector store, while the original images were stored in a document store, allowing retrieval through textual summaries while preserving the original images for answer generation to reduce potential information loss.

\subsubsection{Multimodal RAG}

Figure \ref{fig:mm_rag_overview} provides an overview of our multimodal RAG approaches, where we combined text and image modalities. We reused the two image retrieval methods from the Image-Only RAG setting, adding text retrieval. We explored two configurations within this setup:
  
\paragraph{Multimodal Embeddings and Separate Vector Stores}
  
We embedded images using CLIP and texts using text-embedding-3-small, storing them in separate vector stores. We embedded the query for both modalities and performed a separate similarity search in each store, ensuring both text and image results are retrieved.

\paragraph{Image Summaries and Combined Vector Store}

We converted images into text summaries and embedded these, along with text chunks extracted from the PDF documents, using text-embedding-3-small. Both were stored in a single vector store. A similarity search was then performed to retrieve the most relevant documents, whether text or image, based on the query embedding.
  
\subsubsection{Gold Standard Context Prompting}
  
In the Gold Standard Context setting, we directly provided the annotated context from the test set to the LLM along with the question, skipping the retrieval step. This setup serves as an upper bound, demonstrating the performance achievable with perfect retrieval and enabling a direct comparison between the generating models (GPT-4V and LLaVA) independently of retrieval performance.

\section{Evaluation Framework}
\label{sec:evaluation_framework} 

We evaluate the performance of the RAG pipelines using an LLM-as-a-Judge approach \citep{chiang2024chatbot, Zhang2023GPT4V_evaluator}, developing a custom evaluation framework tailored to multimodal data. This framework incorporates metrics similar to those in existing text-only RAG evaluation frameworks, such as RAGAs \citep{es-etal-2024-ragas}.
However, our framework enables evaluating multimodal content to handle both text and images, ensuring a comprehensive assessment of the RAG system's performance. We make the code for all experiments and the evaluation framework available at the following URL: \url{https://github.com/riedlerm/multimodal_rag_for_industry}.
  
The framework is designed to be modular and can be used with multiple models as evaluators, including GPT-4V and LLaVA. The core of the framework consists of an evaluation module that includes specialized evaluators for each metric. These evaluators construct prompts for the model, execute the evaluation, and parse the model's output to ensure it meets the required format.
The delivered output for each metric includes a binary grade (either 1 or 0) and a reason for the judgment to facilitate easy aggregation and analysis of the results. The final score for each metric is obtained by averaging all binary evaluations over the dataset.

\begin{figure*}[ht!]
    \centering
    \begin{subfigure}{0.30\textwidth}
        \centering
        \includegraphics[width=\textwidth]{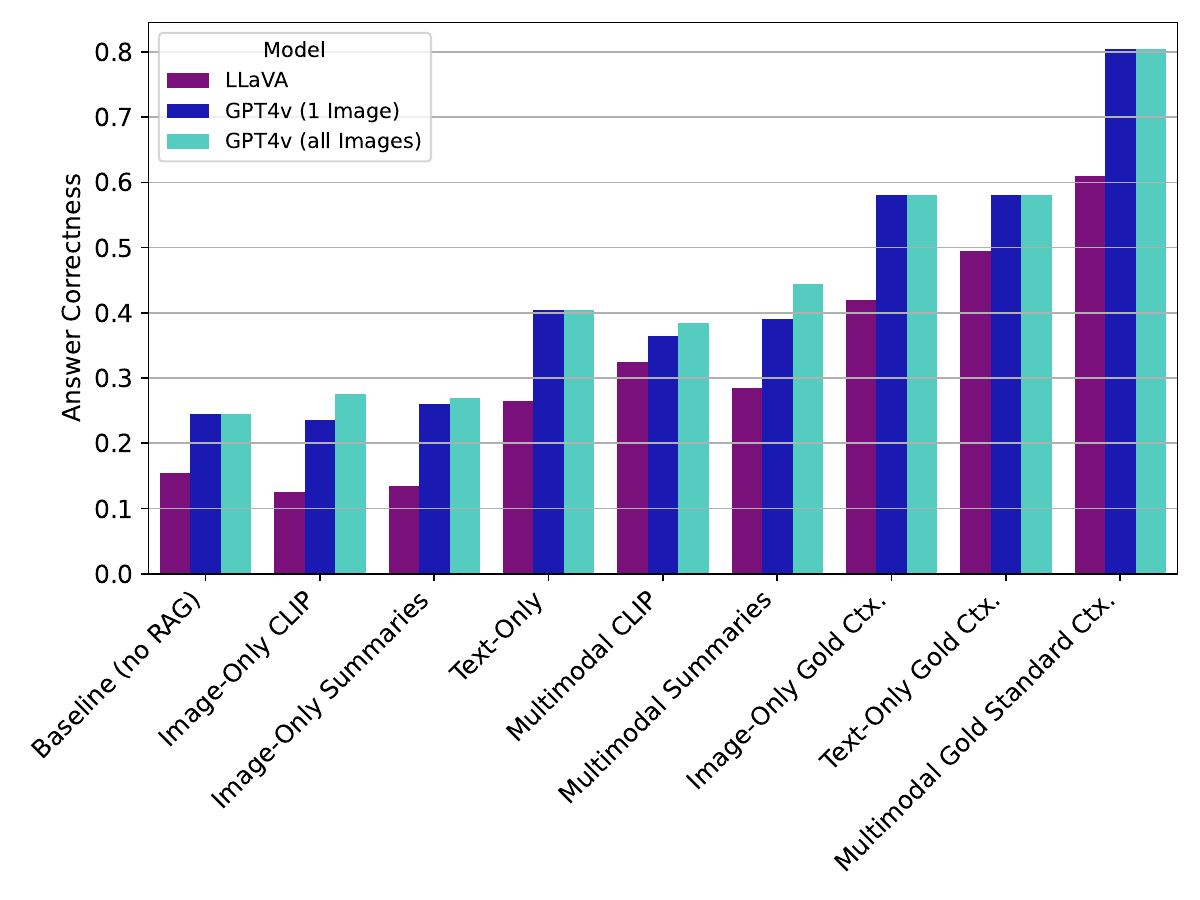}
        \caption{Answer Correctness}
        \label{fig:ans_corr}
    \end{subfigure}
    \begin{subfigure}{0.30\textwidth}
        \centering
        \includegraphics[width=\textwidth]{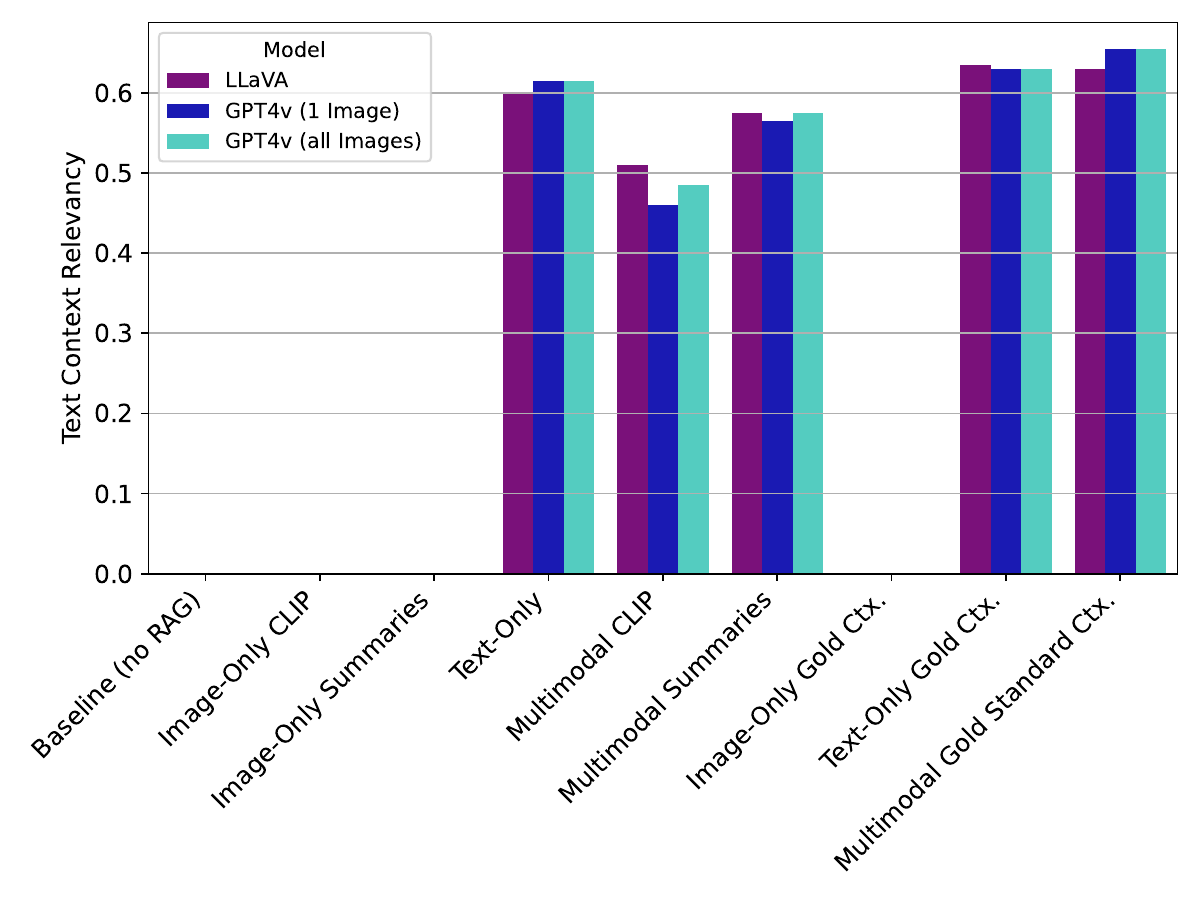}
        \caption{Text Context Relevancy}
    \end{subfigure}
    \begin{subfigure}{0.30\textwidth}
        \centering
        \includegraphics[width=\textwidth]{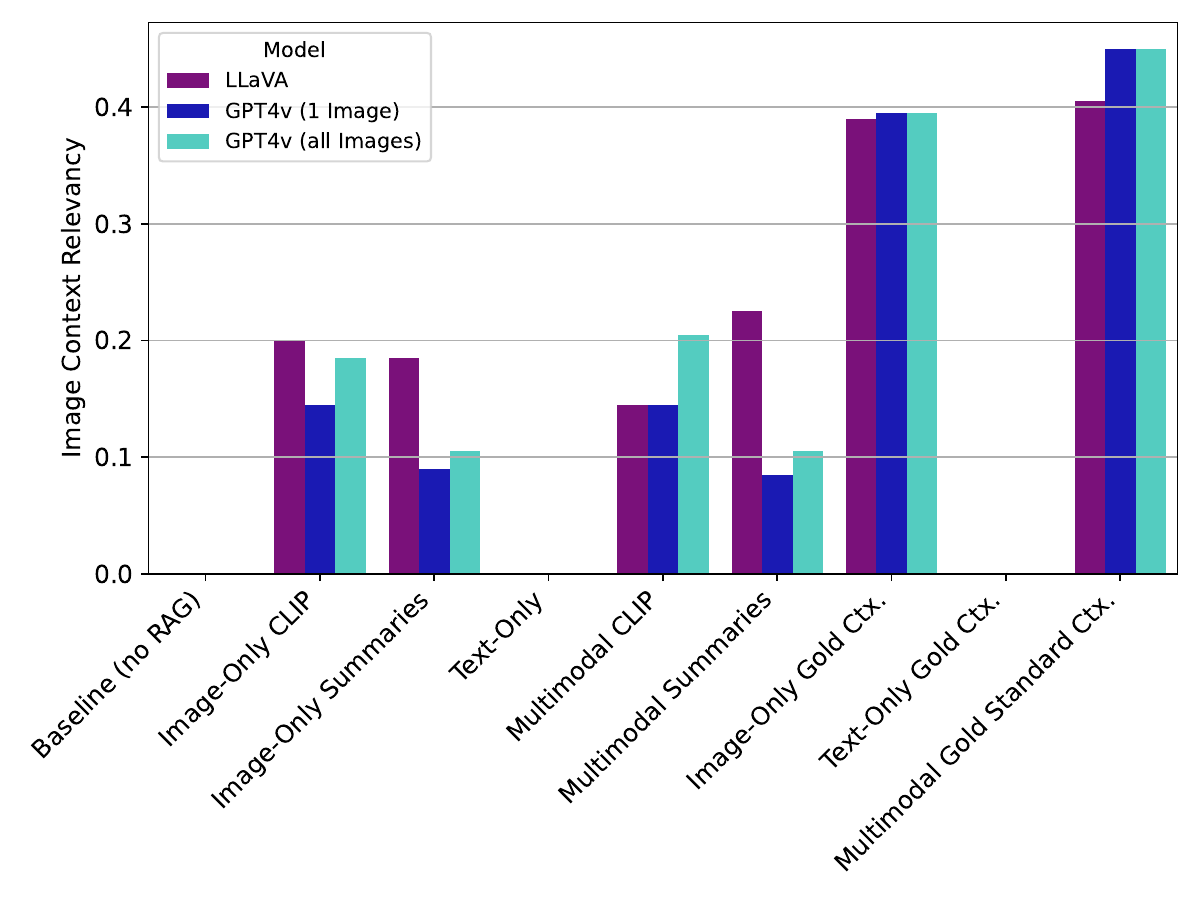}
        \caption{Image Context Relevancy}
    \end{subfigure}

    \vspace{0.5cm}

    \begin{subfigure}{0.30\textwidth}
        \centering
        \includegraphics[width=\textwidth]{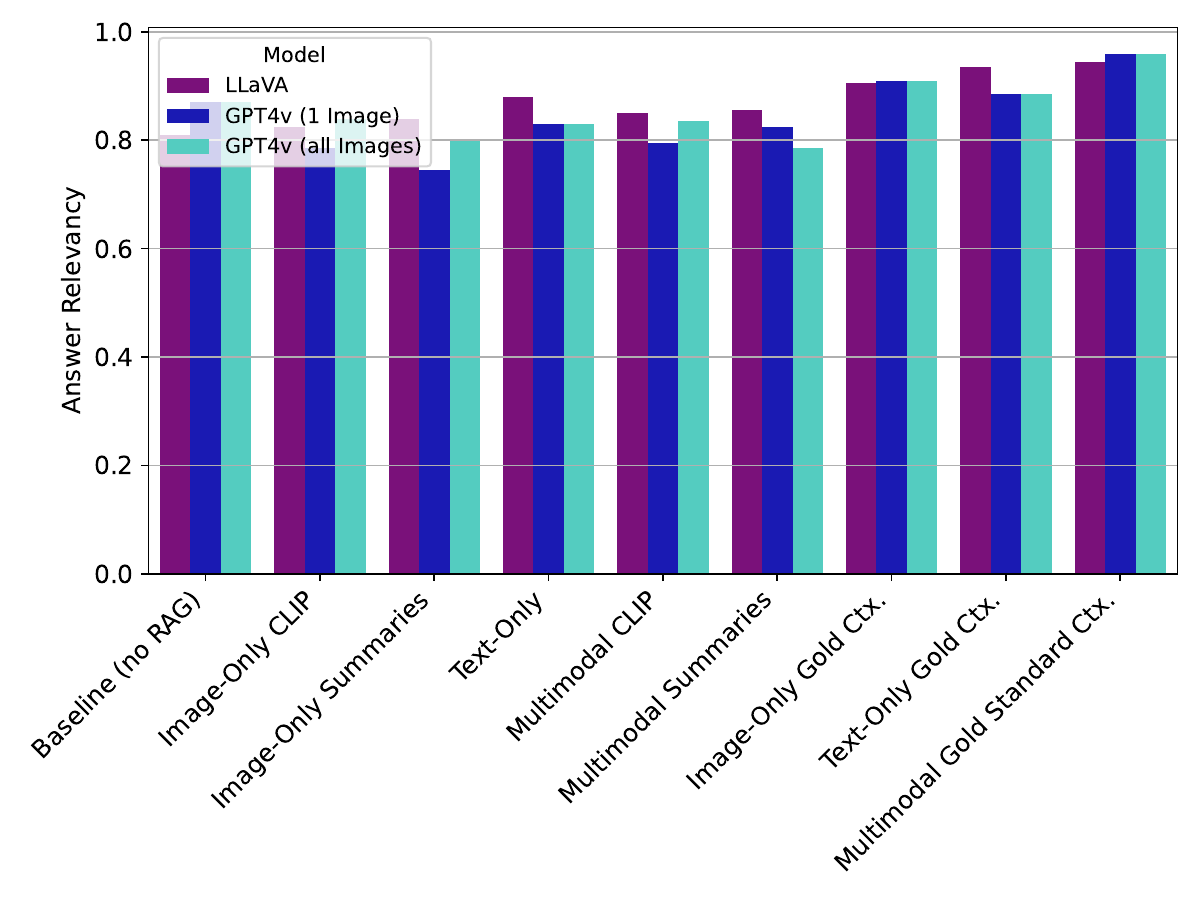}
        \caption{Answer Relevancy}
    \end{subfigure} 
    \begin{subfigure}{0.30\textwidth}
        \centering
        \includegraphics[width=\textwidth]{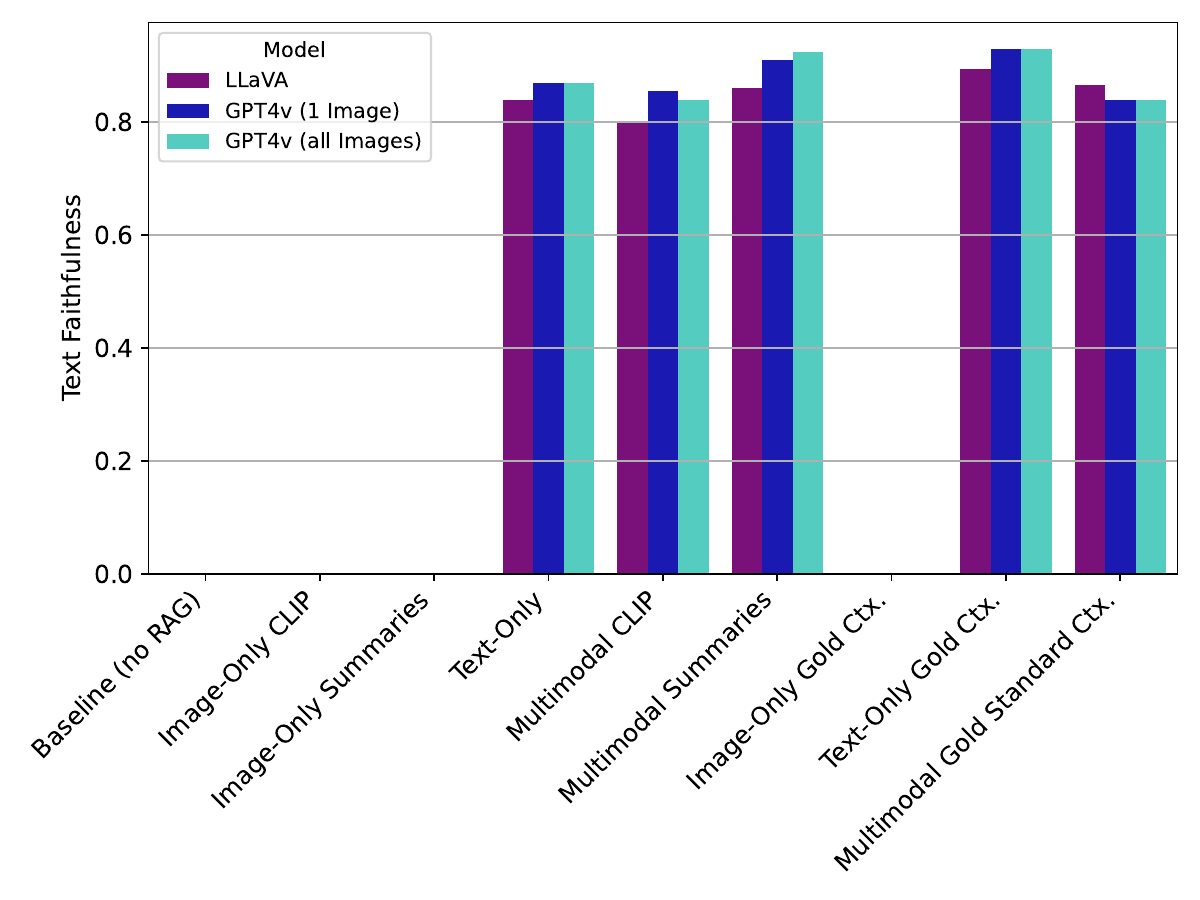}  
        \caption{Text Faithfulness}
    \end{subfigure}
    \begin{subfigure}{0.30\textwidth}
        \centering
        \includegraphics[width=\textwidth]{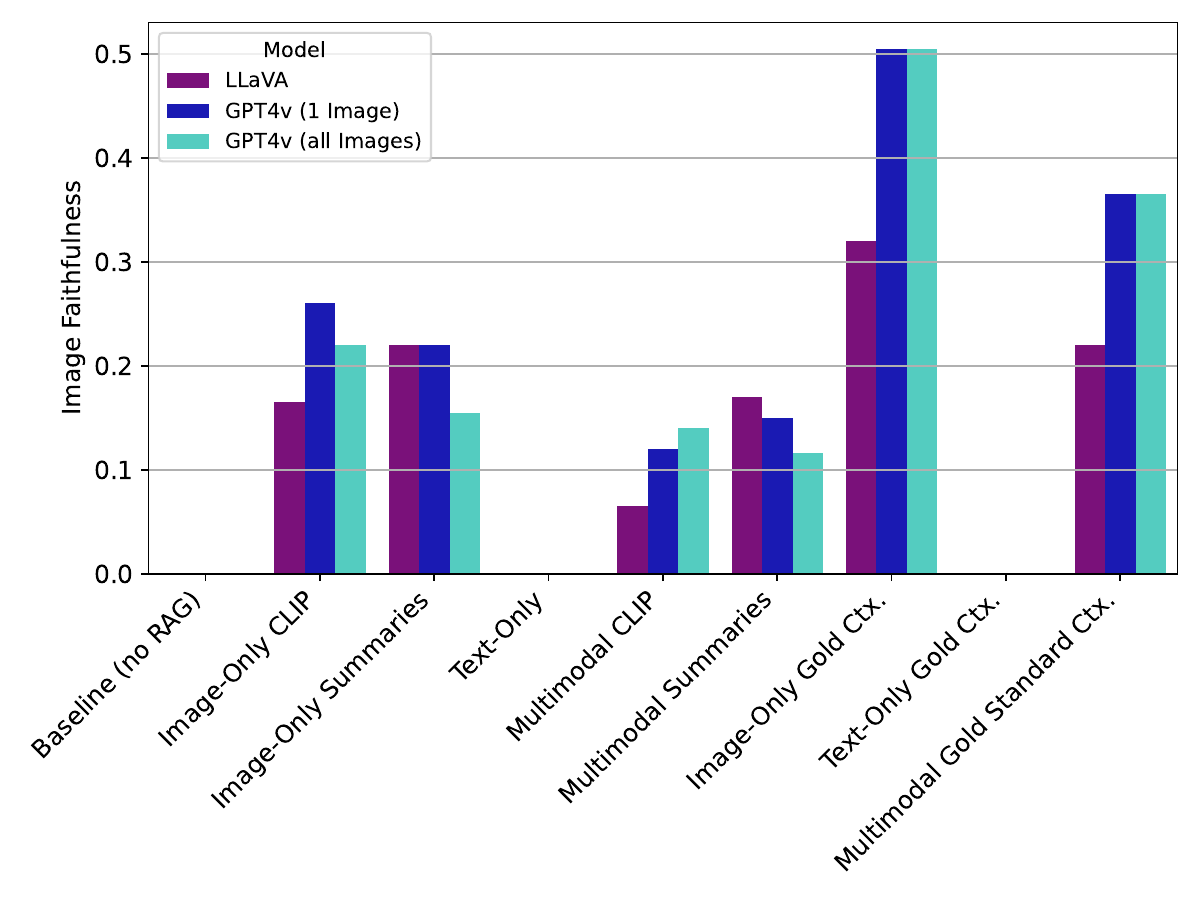}   
        \caption{Image Faithfulness}
    \end{subfigure}

    \caption{RAG evaluation results for GPT-4V prompted with either a single or multiple images, and LLaVA (always single image) across six metrics. The results show the performance of each RAG setting in generating accurate, relevant, and faithful responses based on both text and image inputs.}

    \label{fig:all_scores}
\end{figure*}

\paragraph{Evaluation Metrics}

The framework employs six evaluation metrics: \textbf{Answer Correctness} uses reference-guided pairwise comparison to evaluate the correctness of the generated answer compared to a reference answer and is the only metric relying on the presence of ground truth answers; \textbf{Answer Relevancy} assesses whether the generated answer is relevant to the question; \textbf{Text Faithfulness} measures the consistency between the generated answer and the retrieved textual context; \textbf{Image Faithfulness} evaluates whether the generated answer aligns with the content of the retrieved images; \textbf{Text Context Relevancy} evaluates the relevancy of the retrieved text context in answering the question; and \textbf{Image Context Relevancy}, assesses the relevancy of the retrieved images to the question. We summarize these metrics in Table \ref{tab:evaluation_metrics} and report the prompts to calculate each metric in Appendix \ref{sec:eval_prompt_templates}.

\begin{table}[h!]
    \centering
    \begin{tabular}{ll}

        \hline
        \textbf{Metric} & \textbf{Required Inputs} \\
        \hline
        Ans. Correctness & Q, GA, RA \\
        Ans. Relevancy & Q, GA \\
        Text Faithfulness & GA, Text Ctx. \\
        Img. Faithfulness & GA, Img. Ctx. \\
        Text Ctx. Relevancy & Q, Text Ctx. \\
        Img. Ctx. Relevancy & Q, Img. Ctx. \\
        \hline
    \end{tabular}
    \caption{Evaluation metrics and their required inputs: Q is the question, GA is the generated answer, RA is the reference answer, and Text/Img. Ctx. are the context provided as text/image respectively.}
    \label{tab:evaluation_metrics}
\end{table}

\section{Results}

We summarize the performance of GPT-4V and LLaVA in 9 different settings including single-modality and multimodal RAG approaches in Figure \ref{fig:all_scores}. 
Unlike LLaVA, which was limited to processing a single image per prompt during our experiments\footnote{For our experiments we employed \url{llava-hf/llava-v1.6-mistral-7b-hf}}, GPT-4V has the capability to handle multiple images and interleaved text-image sequences. We assessed its performance both using one image, for comparison with LLaVA, and multiple images to inspect the effect of using multiple images as context.
In settings utilizing image summaries for retrieval, the summarizing model (LLaVA or GPT-4V) is consistently used for answer synthesis.
% We assess our RAG pipelines using the framework described in \ref{sec:evaluation_framework}. 
Both models, i.e., LLaVA and GPT-4V, are used for evaluation. This means that the final score is derived from averaging evaluations conducted with both models, regardless of the generator model chosen. This approach helps mitigate self-enhancement bias \citep{chiang2024chatbot} and avoids single-judge evaluations.
We report the full results of our experiments in Appendix \ref{sec:detailed_results}.

\subsection{Single-Modality vs. Multimodal RAG}

In Figure \ref{fig:ans_corr}, we show the Answer Correctness for single-modality and multimodal settings to investigate whether a combination of text and images improves performance. The upper bound results, obtained by prompting with the gold standard context, reveal that using both text and images significantly outperforms single-modality approaches. This suggests that integrating images with text is beneficial for this domain.
In RAG settings, results show that multimodal RAG can outperform single-modality approaches. Image-only RAG tends to yield the lowest scores, only slightly outperforming the baseline, highlighting the need for improved image retrieval mechanisms. Conversely, multimodal RAG using image summaries slightly outperforms text-only RAG, although the difference is smaller compared to the gold standard context setting. The multimodal setting using CLIP embeddings shows mixed results: LLaVA performs better with multimodal inputs, while GPT-4V performs better with text-only RAG.
Overall, while multimodal RAG offers an advantage over text-only RAG, particularly in scenarios with effective text and image retrieval, image retrieval still requires further improvement.

\subsection{Prompting with Gold Standard Context}

When provided with the gold standard context, using both text and images yields significantly higher Answer Correctness scores compared to single-modality approaches (Figure \ref{fig:ans_corr}). Around 60\% of the questions can be answered using a single modality, given the correct context. However, using both modalities increases the Answer Correctness to approximately 80\%, suggesting that a combination of text and image is often required for a correct answer. Despite these gains, image retrieval appears more challenging compared to text retrieval, as evidenced by the pronounced gap between the single-modality gold context and their respective single-modality RAG settings.

\subsection{Prompting with Multiple Images}

Employing multiple images in prompts generally improves performance across all metrics, except for Image Faithfulness, in both Image-Only and Multimodal RAG settings. This enhancement suggests that additional images increase the chance of incorporating the relevant context in the prompt, thereby improving Answer Correctness and Relevancy. However, the model's tendency to focus on one image might explain lower Image Faithfulness scores. Overall, the results highlight the benefits of processing multiple images within a single prompt.

\subsection{GPT-4V vs. LLaVA}

In our experiments GPT-4V consistently outperforms LLaVA in terms of Answer Correctness, often by a significant margin. The same trend is observed for Text Faithfulness, albeit less pronounced.
The image-related metrics present a mixed picture. While some data points show LLaVA substantially surpassing GPT-4V in Image Context Relevancy, other results suggest the opposite trend. In terms of Answer Relevancy, LLaVA slightly outperforms GPT-4V in most settings.

\subsection{Multimodal Embeddings vs. Image Summaries}

In both Image-Only and Multimodal RAG settings, our two image processing strategies show comparable performance. However, the image summaries setup slightly outperforms multimodal embeddings across most metrics, except for Image Context Relevancy. The image summaries approach appears to be more promising, as it offers greater potential for future advancements. 
For instance, the summarization prompt can be tailored to focus on specific image aspects and include few-shot examples, options not possible with multimodal embeddings. Additionally, the summarization model can be further optimized with task-specific models, and there are more choices for text embedding models compared to multimodal ones. In contrast, the multimodal embedding approach relies heavily on the quality of the embedding model, limiting its potential for improvement.

\section{Conclusion}

Our research demonstrates the potential of integrating multimodal models into RAG systems for the industrial domain. By incorporating both text and images, we observed significant improvements in performance, particularly when the retrieval process successfully identifies relevant texts and images. Leveraging textual summaries from images provides greater flexibility and optimization opportunities compared to multimodal embeddings. Despite the challenges associated with image retrieval, our findings underscore the importance of multimodal data in enhancing the quality of generated answers.
Future work will focus on refining image retrieval, comparing our results with fine-tuning-based approaches, and combining RAG with multimodal LLMs fine-tuned for the industrial domain. Additionally, we plan to evaluate each pipeline step independently on domain-specific data to better identify potential failure points, especially in image retrieval, even though our current analysis focused on end-to-end metrics.

\section{Limitations}
\label{sec:limitations}

While our study demonstrates promising results, several areas require further investigation. First, the lack of publicly available, domain-specific datasets restricts the reproducibility and generalizability of our findings, highlighting the need for future work to develop such resources.

While GPT-4V and LLaVA were effective, they share common LLM limitations, including inaccuracies, hallucinations, and difficulty handling complex multimodal inputs.
Additionally, our evaluation relies on LLMs, which may introduce biases. While we mitigated this by using both GPT-4V and LLaVA, human evaluations remain essential for a reliable assessment.
New multimodal models, such as GPT-4o \citep{openai2024gpt4o}, emerged during our study but could not be included. Continuous evaluations with the latest models are necessary.

Although we focused on the industrial domain, we believe our approach can be applied to other domains as well, since no domain-specific components were used at any stage of our pipeline. Additional experiments can help confirm the applicability of our findings in other fields, such as healthcare or finance.

\bibliography{custom}

\appendix

\section{Implementation Details}
\label{sec:implementation_details}

\subsection{Vector Databases and Retrievers}

For our RAG pipelines, we chose ChromaDB \citep{chromadb} as our vector database because of its open-source nature, local usability, and seamless integration with LangChain \citep{langchain}, which we used as the framework for our experiments. We employed the Hierarchical Navigable Small World (HNSW) search method \citep{hnsw} along with L2 similarity search.

Embeddings were generated using either \texttt{CLIP ViT-L/14} via the OpenCLIP implementation \citep{Cherti2022openclip} for settings with multimodal embeddings, or OpenAI's \texttt{text-embedding-3-small}\footnote{\url{https://platform.openai.com/docs/guides/embeddings}} for text and image summaries.

For retrieval, we employed LangChain's VectorStoreRetriever for CLIP embeddings and the MultiVectorRetriever for text and image summaries embedded with \texttt{text-embedding-3-small}.

\subsection{Number of Retrieved Documents}

We maintained the parameter \( k \), representing the number of documents retrieved during the pipeline's retrieval step, at LangChain's default value of 4 across all experiments to ensure consistency. In settings with separate vector stores and retrievers for text and image embeddings, \( k \) was set to 2 for each modality, ensuring a total of 4 retrieved documents. Consequently, in these settings, the top 2 images and the top 2 texts were always retrieved. Conversely, in settings with a single vector store for both texts and textual summaries, the distribution of retrieved texts and summaries varied depending on the relevance of the documents. This variation implies that if the top 4 retrieved documents were all texts, there could be instances where no images were retrieved, and vice versa.
We acknowledge that this approach might not be ideal, as the fixed value of \( k \) may not always capture the most relevant documents, especially in cases where the relevance distribution between text and images is uneven. However, we adopted this strategy to maintain uniformity and simplicity across all experiments, thereby facilitating a controlled comparison of the retrieval mechanisms.

\subsection{Model Versions and Hyperparameters}

\begin{table}[h!] 
\small
  \centering    
  \begin{tabular}{lcc}  
  \hline
  \textbf{Configuration} & \textbf{GPT-4V} & \textbf{LLaVA} \\  
  \hline  
  version & 2024-02- & llava-v1.6- \\ 
   & 15-preview & mistral-7b \\  
  temperature & 0.7 & 1 \\  
  top\_p & 0.95 & 1 \\  
  max\_tokens & 300 & 300 \\  
  \hline  
  \end{tabular}  
  \caption{Model Versions and Hyperparameters}  
  \label{table:model_hyperparameters}  
\end{table}

\section{Prompt Templates}

\subsection{RAG Prompt Templates}

\begin{figure}[h!]
  \centering
  \includegraphics[width=\columnwidth]{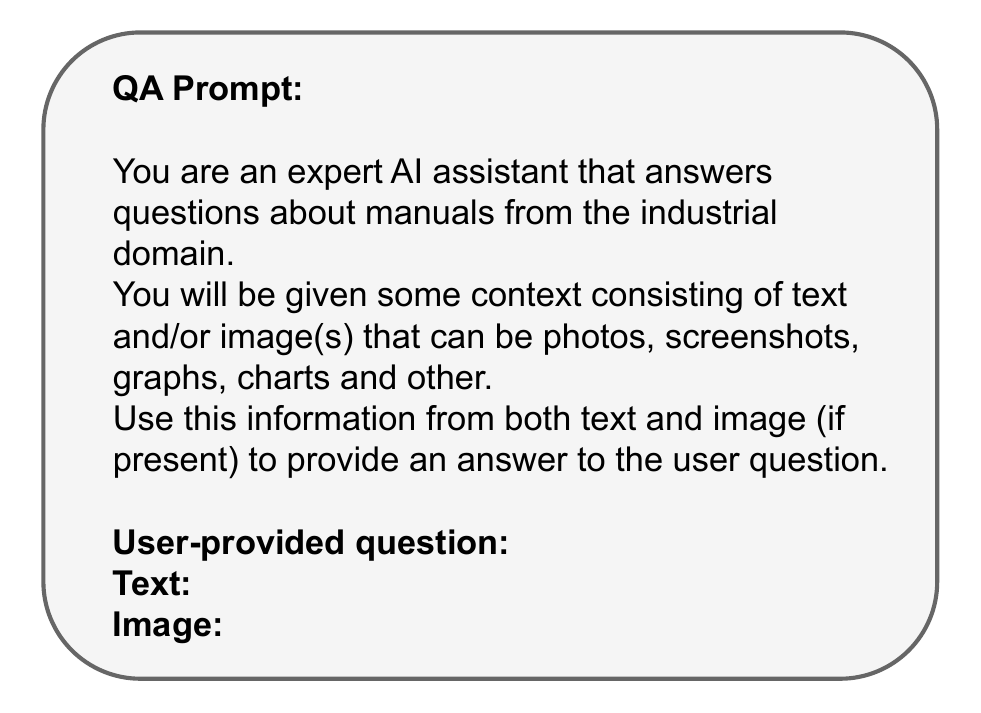}

  \caption{Question Answering Prompt.}
  \label{fig:qa_template.pdf}
\end{figure}

\begin{figure}[h!]

\centering
  \includegraphics[width=\columnwidth]{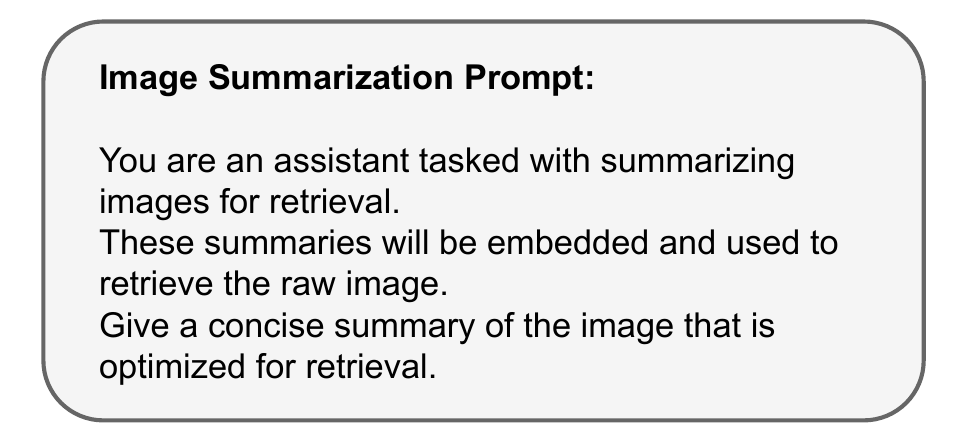}
  \caption{Image Summarization Prompt.}
  \label{fig:summarization_template.pdf}
\end{figure}

\subsection{Evaluation Prompt Templates}
\label{sec:eval_prompt_templates}

\begin{figure}[h!]
  \includegraphics[width=\columnwidth]{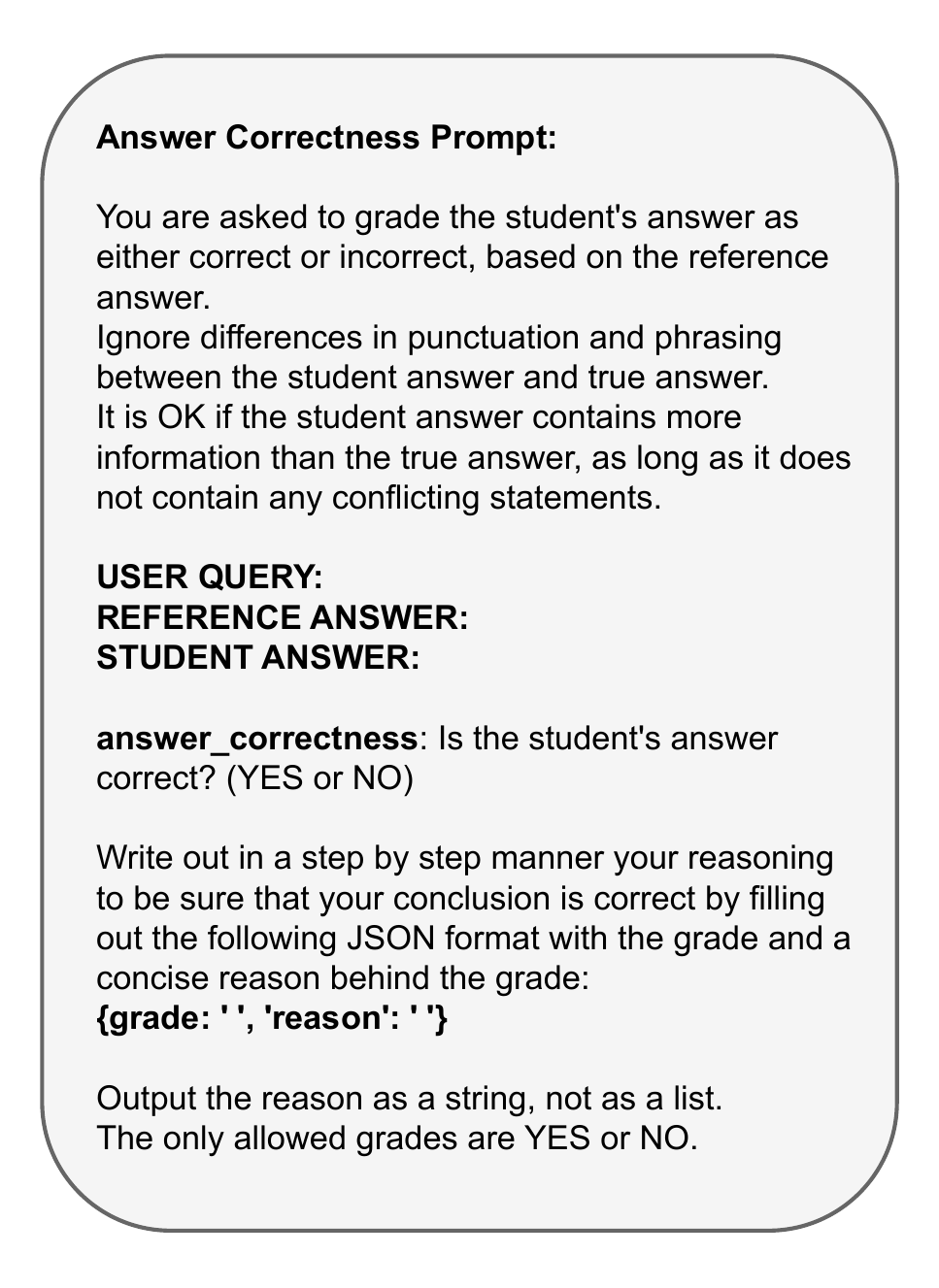}
  \caption{Answer Correctness Evaluation Prompt.}
  \label{fig:ans_corr_template.pdf}
\end{figure}

\begin{figure}[H]
  \includegraphics[width=\columnwidth]{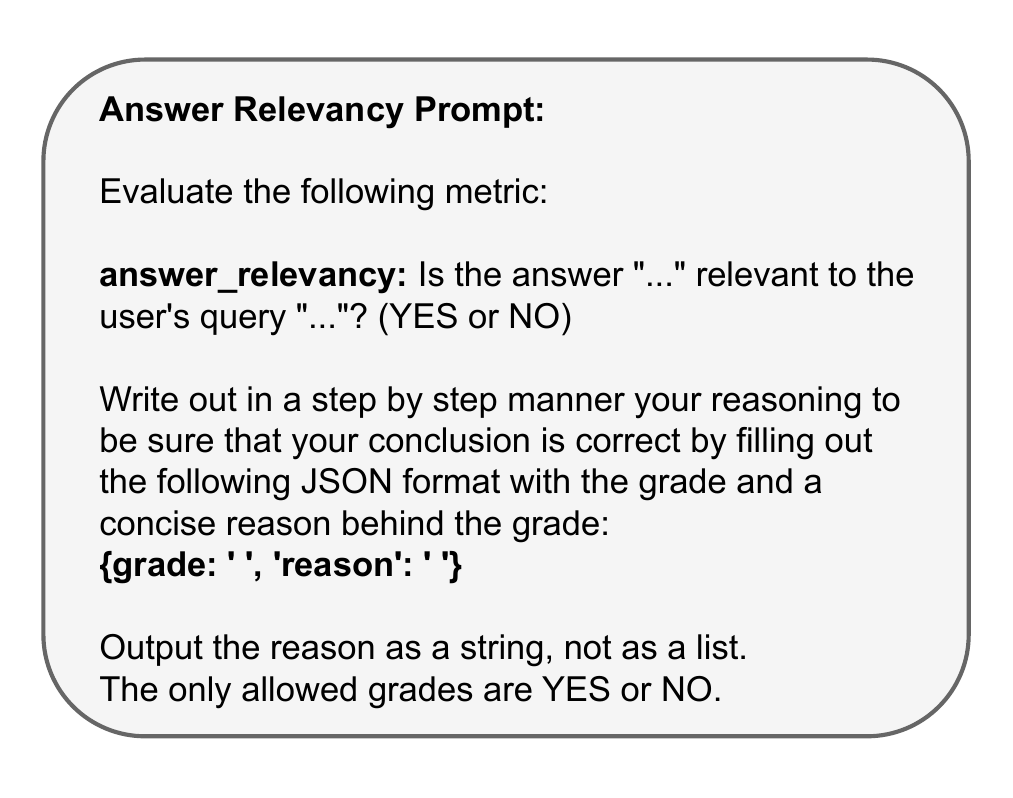}
  \caption{Answer Relevancy Evaluation Prompt.}
  \label{fig:ans_rel_template.pdf}
\end{figure}

\begin{figure}[h!]
  \includegraphics[width=\columnwidth]{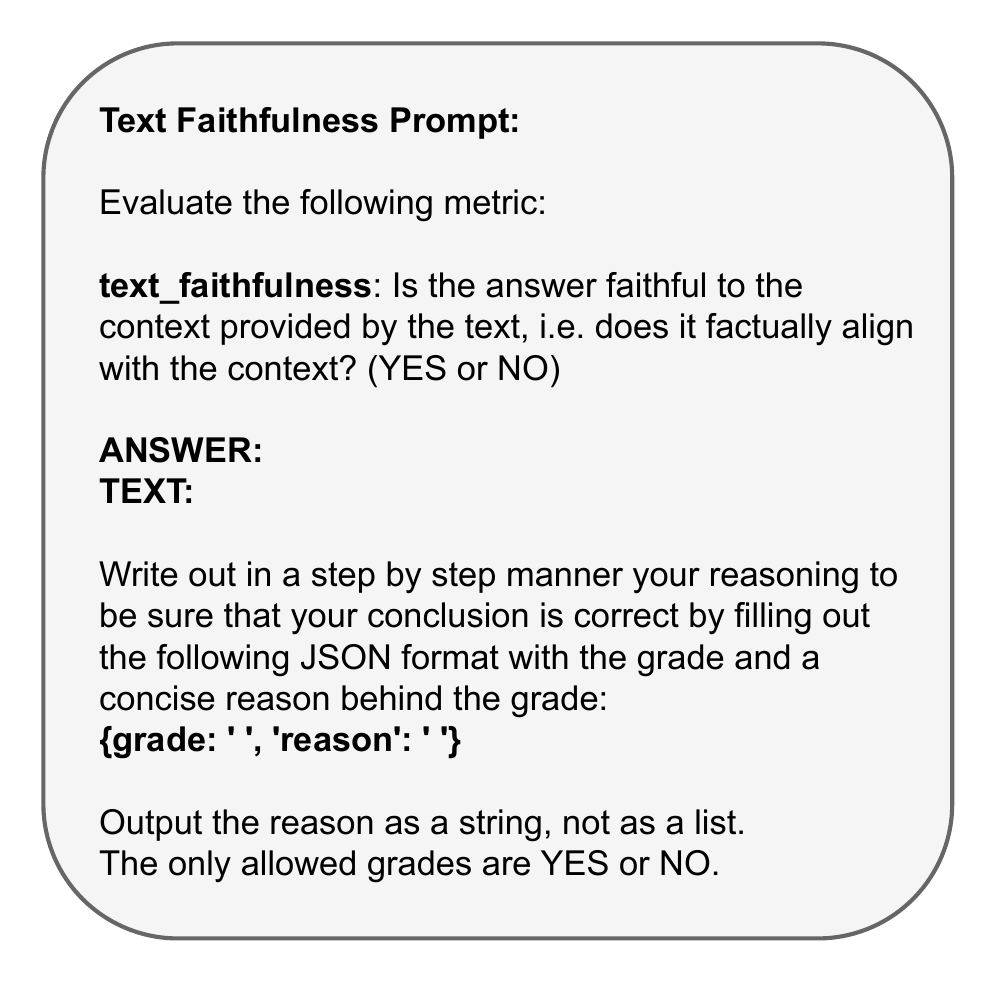}
  \caption{Text Faithfulness Evaluation Prompt.}
  \label{fig:text_faith_template.pdf}
\end{figure}

\clearpage

\begin{figure}[h!]
  \includegraphics[width=\columnwidth]{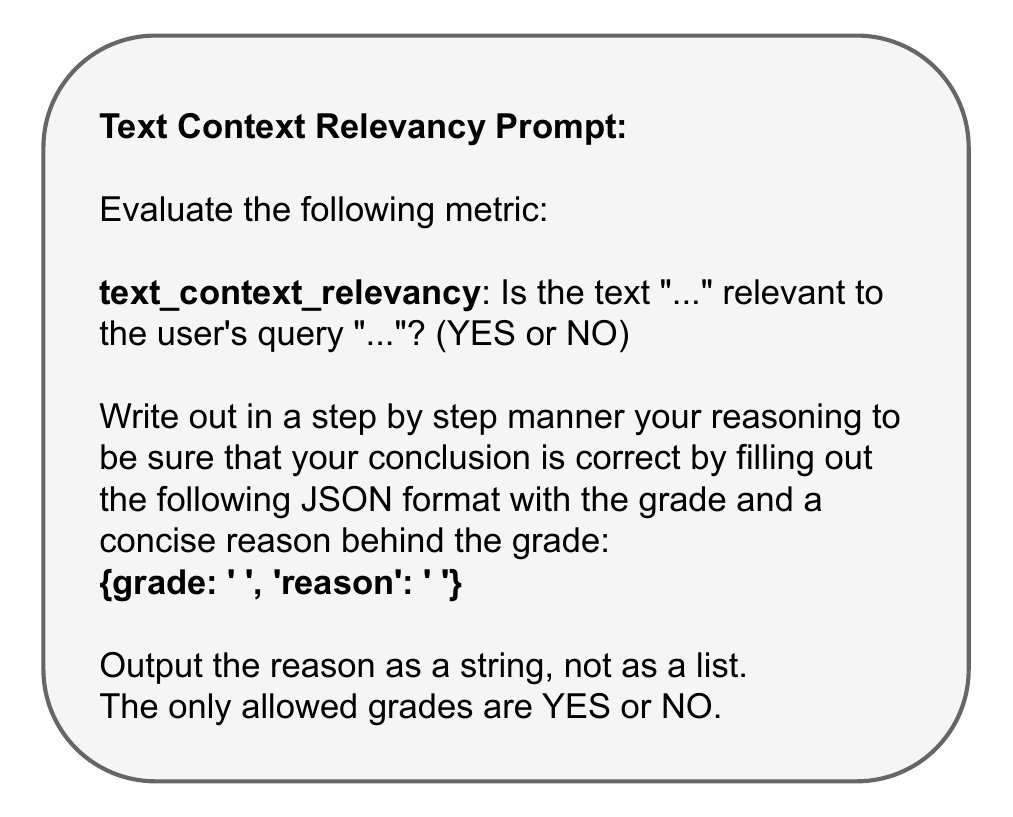}
  \caption{Text Context Relevancy Evaluation Prompt.}
  \label{fig:text_ctx_rel_template.pdf}
\end{figure}

\begin{figure}[h!]
  \includegraphics[width=\columnwidth]{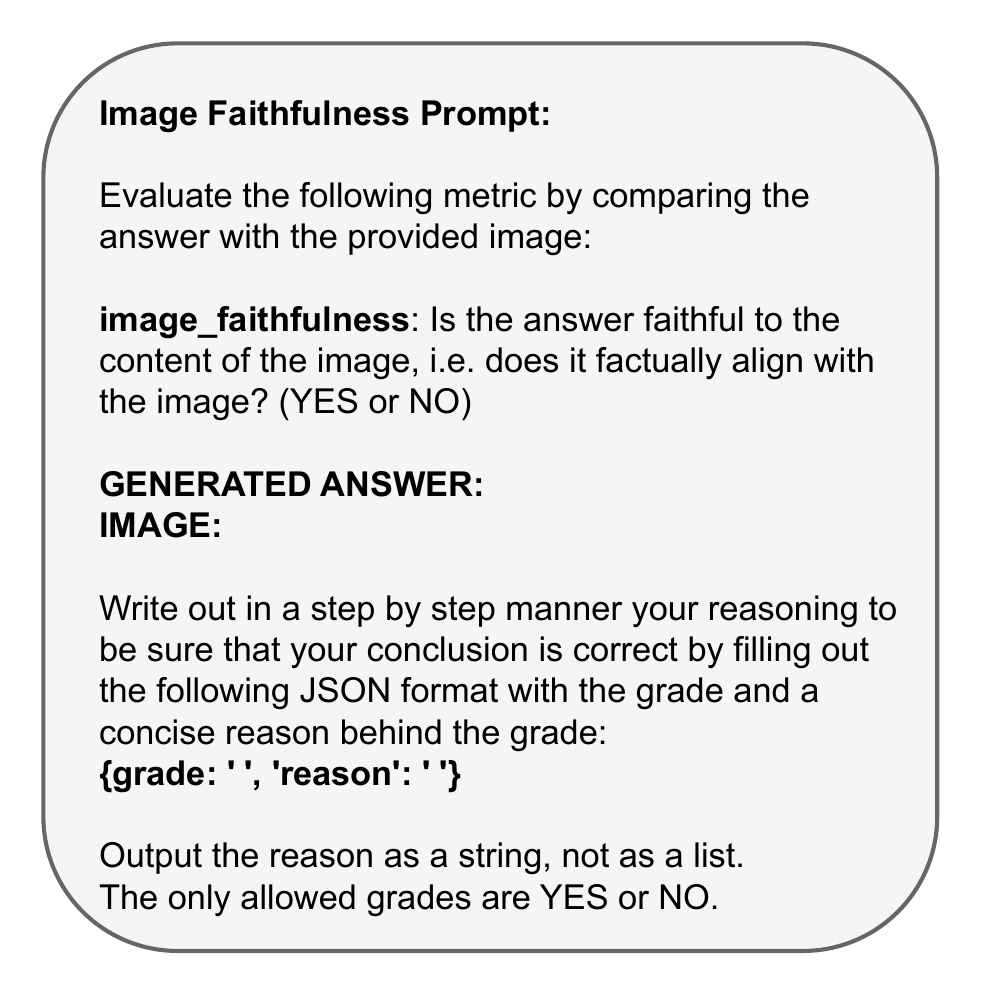}
  \caption{Image Faithfulness Evaluation Prompt.}
  \label{fig:img_faith_template.pdf}
\end{figure}

\begin{figure}[h!]
  \includegraphics[width=\columnwidth]{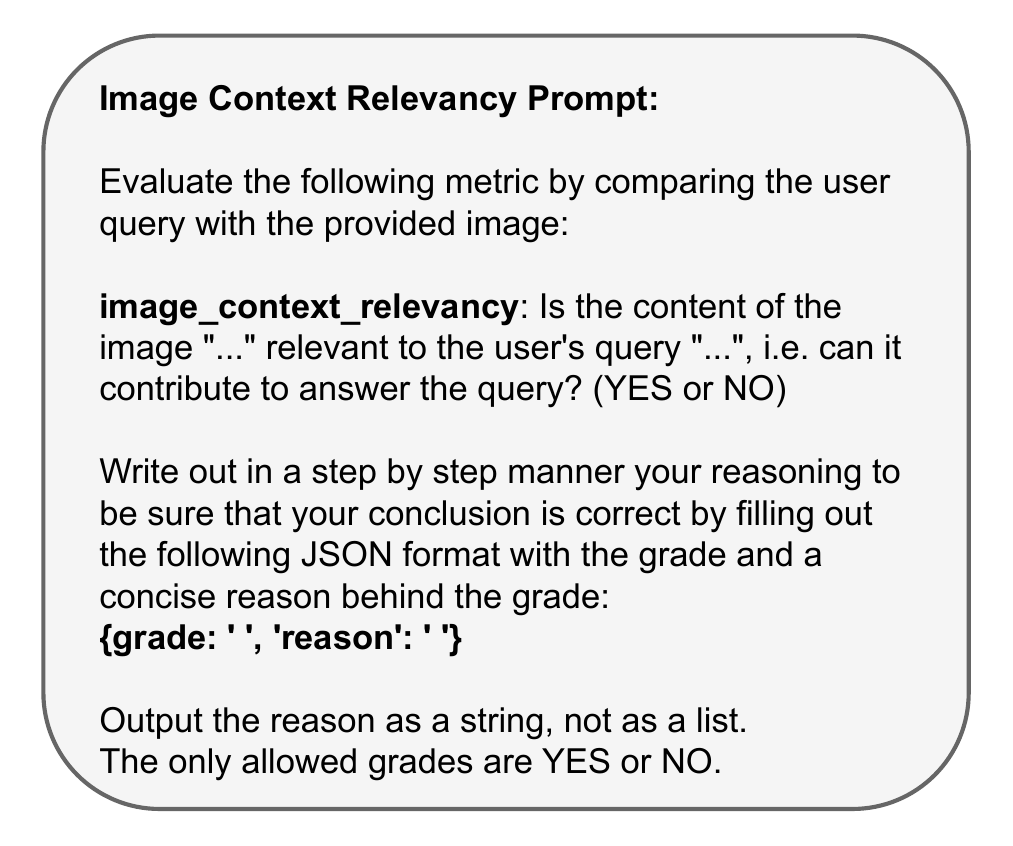}
  \caption{Image Context Relevancy Evaluation Prompt.}
  \label{fig:img_ctx_rel_template.pdf}
\end{figure}

\section{Dataset Details} 
\label{sec:dataset_creation}

\subsection{Text and Image Extraction}

We used the PyMuPDF\footnote{\url{https://github.com/pymupdf/PyMuPDF}} library to extract text and images from industrial domain PDFs, creating a structured dataset stored in a parquet file. Each entry represents a page, with text and corresponding images aligned by page. If a page contained multiple images, each image was stored as a separate entry, along with the text of the page.

\subsection{Question and Answer Annotation}

To generate the RAG test set, we manually annotated 100 question-answer pairs. The questions were designed to reflect typical queries in the industrial domain, such as operational procedures, device configurations, or troubleshooting guidance. These questions were inspired by existing text-only question-context-answer triples from industrial copilot systems, AI assistants that support factory personnel with automation and diagnostics. We introduced the multimodal aspect, incorporating image context to reflect the visual nature of the documents.

The annotation process involved two annotators: one with a base level of understanding of the domain and the other with extensive experience, having worked for many years in the industrial sector, particularly in software development. The annotators used the industrial PDFs as a starting point, manually creating a question based on the content of the document, along with the corresponding answer. For each annotated pair, the annotators also recorded the page number, which was used to extract both the relevant text and image context from the document, forming the quadruples.

\subsection{Example from the Dataset}

\begin{figure*}[h] 
\centering 
\includegraphics[width=\textwidth]{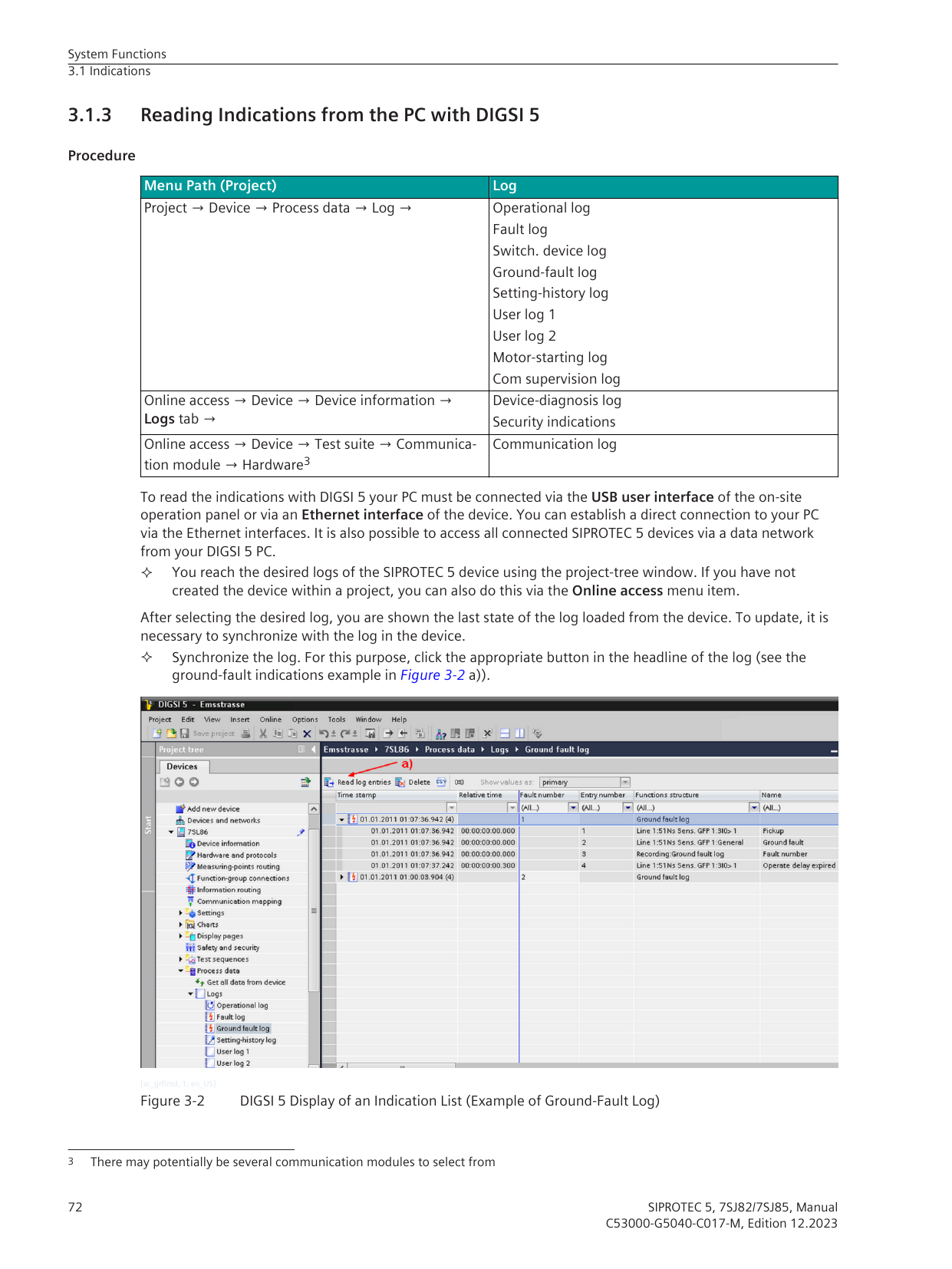} 
\caption{Example page of a PDF file from our document collection.} 
\label{fig:doc_excerpt} 
\end{figure*}

In this section, we present an example from our document collection and RAG test set. First, we show a page from one of the industrial manuals (Figure \ref{fig:doc_excerpt}), followed by the corresponding annotated quadruple, consisting of a question, an answer, the extracted text, and the extracted image (Table \ref{tab:example_qa}).

\begin{table*}[h]
\centering
\begin{tabular}{|p{3cm}|p{10cm}|}
\hline
\textbf{Question} & \small{How can I synchronize the log to read the indications with DIGSI 5?} \\
\hline
\textbf{Answer} & \small{To synchronize the log, click the button `Read log entries' in the headline of the log.} \\
\hline
\textbf{Text Context} & 
\small{
System Functions \newline
3.1 Indications \newline
3.1.3 Reading Indications from the PC with DIGSI 5 Procedure \newline
Menu Path (Project) \newline
Log \newline
Project → Device → Process data → Log → Operational log Fault log \newline
Switch. device log Ground-fault log Setting-history log User log 1 \newline
User log 2 Motor-starting log Com supervision log \newline
Online access → Device → Device information → Logs tab → \newline
Device-diagnosis log Security indications \newline
Online access → Device → Test suite → Communication module → Hardware \newline
Communication log \newline
To read the indications with DIGSI 5 your PC must be connected via the USB user interface of the on-site operation panel or via an Ethernet interface of the device. You can establish a direct connection to your PC via the Ethernet interfaces. It is also possible to access all connected SIPROTEC 5 devices via a data network from your DIGSI 5 PC. \newline
You reach the desired logs of the SIPROTEC 5 device using the project-tree window. If you have not created the device within a project, you can also do this via the Online access menu item. \newline
After selecting the desired log, you are shown the last state of the log loaded from the device. To update, it is necessary to synchronize with the log in the device. \newline
\textcolor{blue}{Synchronize the log. For this purpose, click the appropriate button in the headline of the log (see the ground-fault indications example in Figure 3-2 a)).} \newline
Figure 3-2 DIGSI 5 Display of an Indication List (Example of Ground-Fault Log) \newline
There may potentially be several communication modules to select from \newline
SIPROTEC 5, 7SJ82/7SJ85, Manual C53000-G5040-C017-M, Edition 12.2023
} \\
\hline
\textbf{Image Context} & \includegraphics[width=0.6\textwidth]{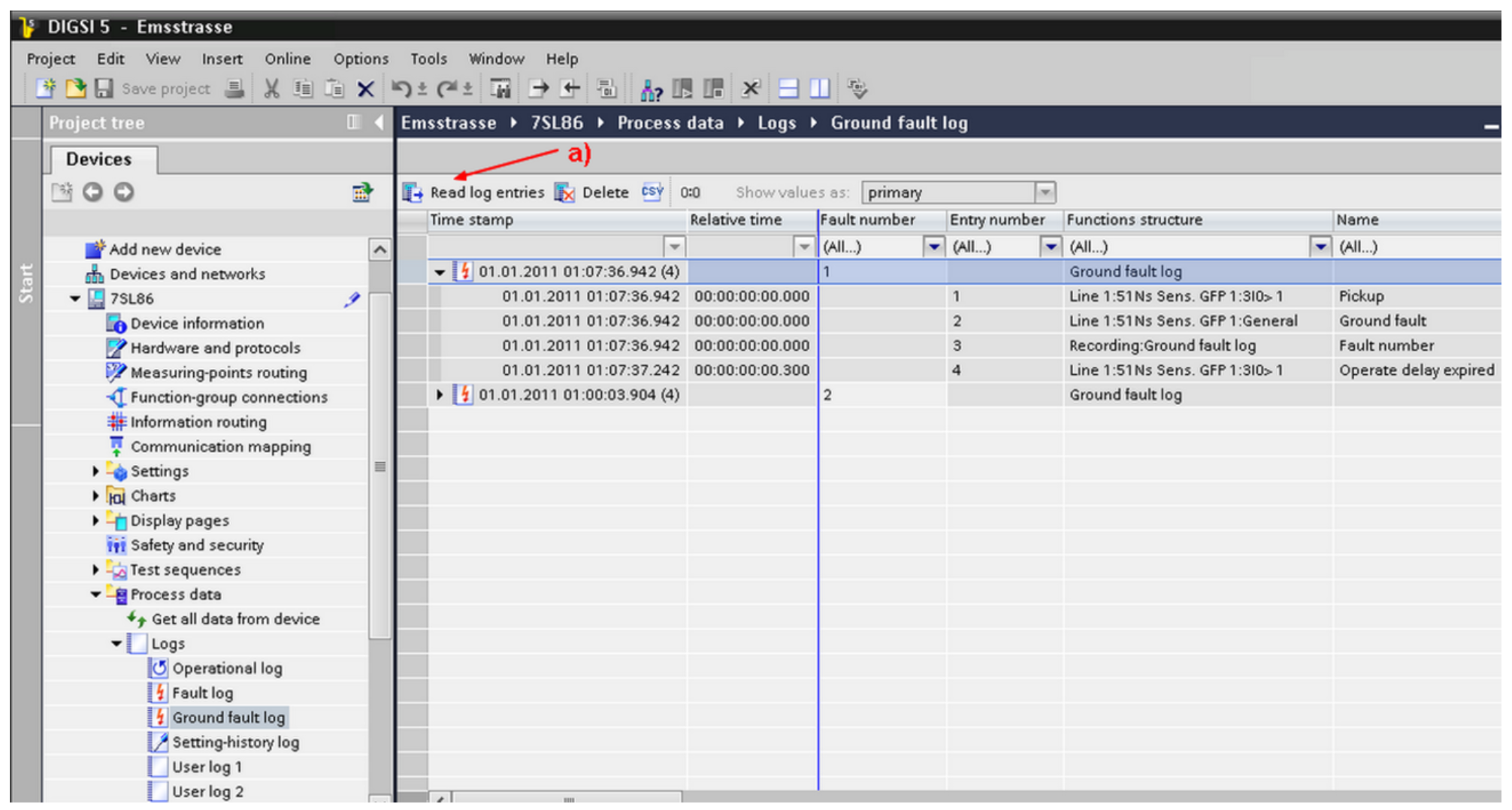} \\
\hline
\end{tabular}
\caption{An example of a multimodal quadruple, incorporating both text and image context. The text highlighted in blue provides useful information to answer the question; however, it is insufficient to identify the `Read log entries' button, which is highlighted in the image and also required for a correct answer.}
\label{tab:example_qa}
\end{table*}

\clearpage
\onecolumn

\section{Detailed Results}
\label{sec:detailed_results}

\begin{table*}[h]
\scriptsize
\centering
\begin{tabular}{lllrrrrrr}
\hline
\textbf{Approach} & \textbf{Generator} & \textbf{Evaluator} & \textbf{Ans. Corr.} & \textbf{Ans. Rel.} & \textbf{Text Faith.} & 
 \textbf{Text Ctx. Rel.} & \textbf{Img. Faith.} & \textbf{Img. Ctx. Rel.} \\

\hline
Baseline & GPT-4V & GPT-4V & 0.18 & 0.96 & -- & -- & -- & --  \\ 
 & GPT-4V & LLaVA & 0.31 & 0.78 & -- & -- & -- & --  \\ 
 & LLaVA & GPT-4V & 0.12 & 0.92 & -- & -- & -- & --  \\
 & LLaVA & LLaVA & 0.19 & 0.70 & -- & -- & -- & --  \\  
\hline

Text-Only & GPT-4V & GPT-4V & 0.39 & 0.91 & 0.76 & 0.63 & -- & -- \\  
RAG & GPT-4V & LLaVA & 0.42 & 0.75 & 0.98 & 0.60 & -- & -- \\ 
 & LLaVA & GPT-4V & 0.29 & 0.95 & 0.69 & 0.63 & -- & -- \\ 
 & LLaVA & LLaVA & 0.24 & 0.81 & 0.99 & 0.57 & -- & -- \\

\hline
Image-Only & GPT-4V MI & GPT-4V & 0.24 & 0.89 & -- & -- & 0.20 & 0.36 \\  
RAG & GPT-4V MI & LLaVA & 0.31 & 0.79 & -- & -- & 0.24 & 0.01  \\
Clip & GPT-4V SI & GPT-4V & 0.19 & 0.86 & -- & -- & 0.32 & 0.28  \\
 & GPT-4V SI & LLaVA & 0.28 & 0.71 & -- & -- & 0.20 & 0.01  \\
 & LLaVA & GPT-4V & 0.11 & 0.82 & -- & -- & 0.04 & 0.39  \\
 & LLaVA & LLaVA & 0.14 & 0.83 & -- & -- & 0.29 & 0.01  \\ 

\hline
Image-Only & GPT-4V MI & GPT-4V & 0.22 & 0.85 & -- & -- & 0.20 & 0.20  \\
RAG & GPT-4V MI & LLaVA & 0.32 & 0.75 & -- & -- & 0.11 & 0.01  \\
Summaries & GPT-4V SI & GPT-4V & 0.21 & 0.80 & -- & -- & 0.30 & 0.18  \\
 & GPT-4V SI & LLaVA & 0.31 & 0.69 & -- & -- & 0.14 & 0.00  \\
 & LLaVA & GPT-4V & 0.10 & 0.86 & -- & -- & 0.11 & 0.36  \\
 & LLaVA & LLaVA & 0.17 & 0.82 & -- & -- & 0.33 & 0.01  \\

\hline 
Multimodal & GPT-4V MI & GPT-4V & 0.37 & 0.91 & 0.14 & 0.40 & 0.74 & 0.59  \\
RAG Clip & GPT-4V MI & LLaVA & 0.40 & 0.76 & 0.14 & 0.01 & 0.94 & 0.38  \\
 & GPT-4V SI & GPT-4V & 0.35 & 0.86 & 0.11 & 0.29 & 0.76 & 0.57  \\
 & GPT-4V SI & LLaVA & 0.38 & 0.73 & 0.13 & 0.00 & 0.95 & 0.35  \\
 & LLaVA & GPT-4V & 0.30 & 0.94 & 0.05 & 0.28 & 0.62 & 0.63  \\
 & LLaVA & LLaVA & 0.35 & 0.76 & 0.08 & 0.01 & 0.98 & 0.39  \\
\hline

Multimodal & GPT-4V MI & GPT-4V & 0.43 & 0.86 & 0.17 & 0.21 & 0.88 & 0.72  \\
RAG & GPT-4V MI & LLaVA & 0.46 & 0.71 & 0.06 & 0.00 & 0.97 & 0.43  \\
Summaries & GPT-4V SI & GPT-4V & 0.40 & 0.89 & 0.17 & 0.17 & 0.86 & 0.69  \\
 & GPT-4V SI & LLaVA & 0.38 & 0.76 & 0.13 & 0.00 & 0.96 & 0.44  \\
 & LLaVA & GPT-4V & 0.29 & 0.91 & 0.05 & 0.45 & 0.73 & 0.65  \\
 & LLaVA & LLaVA & 0.28 & 0.80 & 0.29 & 0.00 & 0.99 & 0.50  \\

\hline
Text-Only & GPT-4V & GPT-4V & 0.57 & 0.93 & 0.93 & 0.84 & -- & -- \\ 
GSC & GPT-4V & LLaVA & 0.59 & 0.84 & 0.93 & 0.42 & -- & -- \\ 
 & LLaVA & GPT-4V & 0.47 & 0.97 & 0.84 & 0.85 & -- & -- \\ 
 & LLaVA & LLaVA & 0.52 & 0.90 & 0.95 & 0.42 & -- & -- \\ 
\hline

Image-Only & GPT-4V & GPT-4V & 0.53 & 0.94 & -- & -- & 0.68 & 0.75  \\
GSC & GPT-4V & LLaVA & 0.63 & 0.88 & -- & -- & 0.33 & 0.04  \\
 & LLaVA & GPT-4V & 0.21 & 0.95 & -- & -- & 0.24 & 0.74  \\
 & LLaVA & LLaVA & 0.63 & 0.86 & -- & -- & 0.40 & 0.04  \\
\hline

Multimodal & GPT-4V & GPT-4V & 0.78 & 1.00 & 0.50 & 0.86 & 0.78 & 0.89  \\
GSC & GPT-4V & LLaVA & 0.83 & 0.92 & 0.23 & 0.04 & 0.90 & 0.42  \\
 & LLaVA & GPT-4V & 0.59 & 1.00 & 0.26 & 0.77 & 0.80 & 0.85  \\
 & LLaVA & LLaVA & 0.63 & 0.89 & 0.18 & 0.04 & 0.93 & 0.41  \\

\hline
\end{tabular}
\caption{Full evaluation scores across all experiments and metrics. Each setup was run with LLaVA, GPT-4V prompted with a single image (GPT-4V SI), and GPT-4V prompted with multiple images (GPT-4V MI). The quality of the generated textual answer was evaluated with both LLaVA and GPT-4V. GSC refers to prompting with the gold standard context.}
\label{tab:full_results}
\end{table*}

\end{document}